\documentclass{article}

\usepackage{hyperref}       
\usepackage{url}            
\usepackage{booktabs}       
\usepackage{amsfonts}       
\usepackage{nicefrac}       
\usepackage{microtype}      

\usepackage{amsmath}
\usepackage{amssymb}
\usepackage{graphicx}

\usepackage{amsfonts}       

\usepackage{bbm}
\usepackage{bm}

\usepackage{algorithm}
\usepackage{algorithmic}

\usepackage{color}

\newcommand{\mb}[1]{\mathbf{#1}}

\newcommand{\mt}[1]{\mathtt{#1}}
\newcommand{\mc}[1]{\mathcal{#1}}
\renewcommand{\S}{\mathbf{S}}

\newcommand{\ls}[1]  
   {\dimen0=\fontdimen6\the=#1\dimen0
    \advance\lineskip.5\fontdimen5\the\lineskip-\dimen0
    \lineskiplimit=.9\lineskip
    \baselineskip=\lineskip
    \advance\baselineskip\dimen0
    \normallineskip\lineskip
    \normallineskiplimit\lineskiplimit
    \normalbaselineskip\baselineskip
    \ignorespaces
   }

 \date{}

\begin{document}
%
%
\title{A Neural-Symbolic Approach to Design of CAPTCHA}
\author{Qiuyuan Huang, Paul Smolensky, Xiaodong He, Li Deng, Dapeng
Wu\\
\em{\{qihua,psmo,xiaohe\}@microsoft.com, l.deng@ieee.org,
dpwu@ufl.edu}\\ Microsoft Research AI, Redmond, WA
\thanks{This work was carried out while PS was on leave from Johns
Hopkins University. LD is currently at Citadel.  DW is with
University of Florida, Gainesville, FL 32611.} } \maketitle

\begin{abstract}
CAPTCHAs based on reading text are susceptible to
machine-learning-based attacks \cite{CAPTCHA_weblink} due to
recent significant advances in deep learning (DL). To address
this, this paper promotes image/visual captioning based CAPTCHAs,
which is robust against machine-learning-based attacks. To develop
image/visual-captioning-based CAPTCHAs, this paper proposes a new
image captioning architecture by exploiting tensor product
representations (TPR), a structured neural-symbolic framework
developed in cognitive science over the past 20 years, with the
aim of integrating DL with explicit language structures and rules.
We call it the \emph{Tensor Product Generation Network}
(\textbf{TPGN}). The key ideas of TPGN are: 1) unsupervised
learning of \emph{role-unbinding vectors} of words via a TPR-based
deep neural network, and 2) integration of TPR with typical DL
architectures including Long Short-Term Memory (LSTM) models. The
novelty of our approach lies in its ability to generate a sentence
and extract partial grammatical structure of the sentence by using
role-unbinding vectors, which are obtained in an unsupervised
manner. Experimental results demonstrate the effectiveness of the
proposed approach.
\end{abstract}

\ls{1.3}
\section{Introduction}
\begin{figure*}[hbt]
    \centering
    \includegraphics[width=0.45\textwidth]{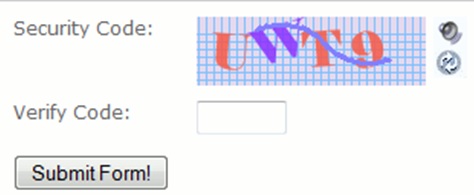}
    \caption{Reading-text-based CAPTCHA.}
    \label{fig:verfication}
\end{figure*}

CAPTCHA, which stands for ``Completely Automated Public Turing
test to tell Computers and Humans Apart", is a type of
challenge-response test used in computers to determine whether or
not the user is a human \cite{CAPTCHA_weblink}.  Most CAPTCHA
systems are based on reading text contained in an image as shown
in Fig.~\ref{fig:verfication}. However, such systems can be easily
cracked by deep learning since deep learning can achieve very high
accuracy in recognizing text. To address this, this paper proposes
a new CAPTCHA, which distinguishes a human from a computer by
testing the capability of image captioning (which generates a
caption for a given image) or video storytelling (which generates
a story consisting of multiple sentences for a give video
sequence). It is known that the performance of computerized image
captioning or video storytelling is far from human performance
\cite{COCO_weblink}. Hence, image/visual captioning based CAPTCHAs
can more reliably distinguish computers from humans.

Deep learning is an important tool in many current natural
language processing (NLP) applications. However, language rules or
structures cannot be explicitly represented in deep learning
architectures. The tensor product representation developed in
\cite{smolensky1990tensor,smolensky2006harmonic} has the potential
of integrating deep learning with explicit rules (such as logical
rules, grammar rules, or rules that summarize  real-world
knowledge). This paper develops a TPR approach for
image/visual-captioning-based CAPTCHAs, introducing the
\emph{Tensor Product Generation Network (\textbf{TPGN})}
architecture.

A TPGN model generates natural language descriptions via learned
representations. The representations learned  in a crucial layer
of the TPGN can be interpreted as encoding grammatical roles for
the words being generated. This layer corresponds to the
role-encoding component of a general, independently-developed
architecture for neural computation of symbolic functions,
including the generation of linguistic structures. The key to this
architecture is the notion of \emph{Tensor Product Representation
(\textbf{TPR})}, in which vectors embedding symbols (e.g., {\tt
lives}, {\tt frodo}) are bound to vectors embedding structural
roles (e.g., {\tt verb}, {\tt subject}) and combined to generate
vectors embedding symbol structures ({\tt [frodo lives]}). TPRs
provide the representational foundations for a general
computational architecture called \emph{Gradient Symbolic
Computation (\textbf{GSC})}, and applying GSC to the task of
natural language generation yields the specialized architecture
defining the model presented here. The generality of GSC means
that the results reported here have implications well beyond the
particular task we address here.

In our proposed image/visual captioning based CAPTCHA, a TPGN
takes an image ${\mathbf I}$ as input and generates a caption.
Then, an evaluator will evaluate the TPGN's captioning performance
by calculating the metric of SPICE \cite{anderson2016spice} by
comparing to human-generated gold-standard captions. If the SPICE
value is less than a threshold, this input image ${\mathbf I}$ can
be used as a challenge in our CAPTCHA.  In this way, we can obtain
a set ${\mathbf B}$ of images to be used as challenges in CAPTCHA.

\begin{figure*}[hbt]
    \centering
    \includegraphics[width=0.7\textwidth]{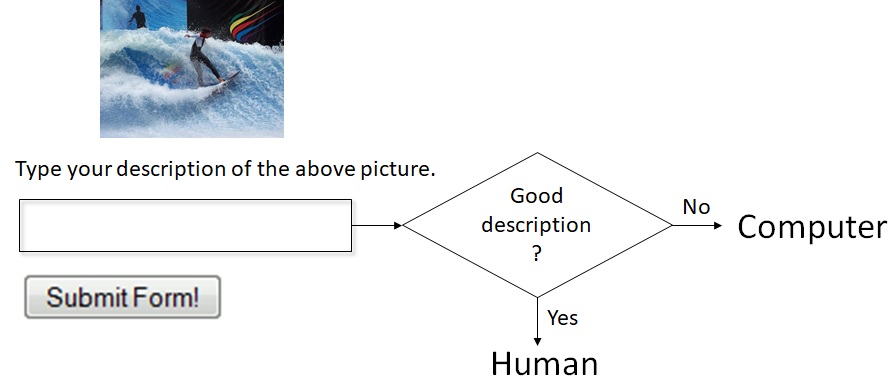}
    \caption{Image-captioning-based CAPTCHA.}
    \label{fig:VerificationCaptioning}
\end{figure*}

In our CAPTCHA shown in Fig.~\ref{fig:VerificationCaptioning}, an
image is randomly selected from the set ${\mathbf B}$ and rendered
as a challenge; a tester is asked to give a description of the
image as an answer.  An evaluator calculates a SPICE value for the
answer.  If the SPICE value is greater than a threshold, the
answer is considered to be correct and the tester is deemed to be
a human; otherwise, the tester is deemed to be a computer.


 The paper is organized as follows.
Section~\ref{sec:design} presents the rationale for our proposed
architecture. In Section~\ref{sec:ExperimentalResults}, we present
our experimental results. Finally, Section~\ref{sec:Conclusion}
concludes the paper.

\begin{figure*}[tbh]
    \centering
    \includegraphics[width=1\textwidth]{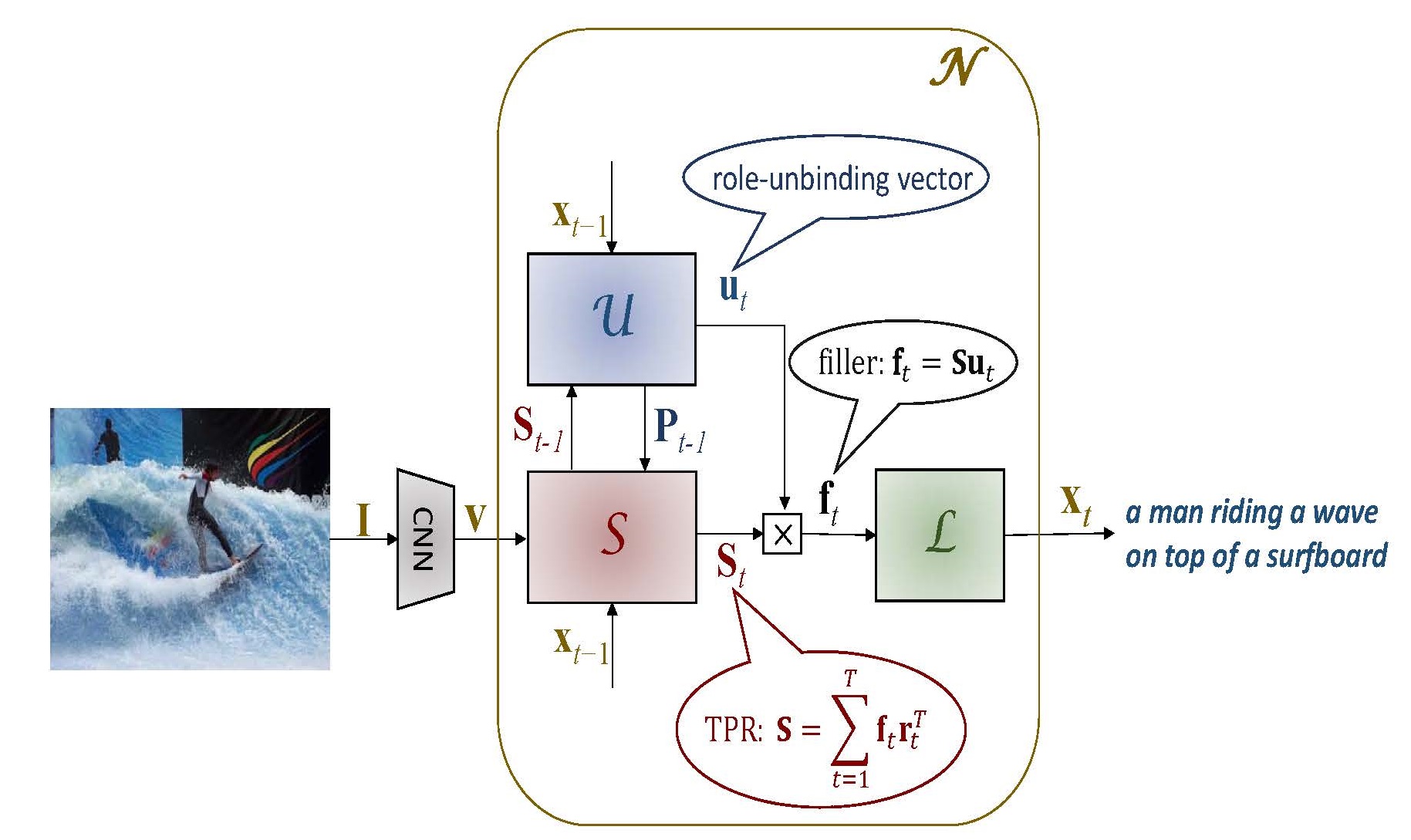}
    \caption{Architecture of TPGN, a TPR-capable generation network. ``$\Box\hspace{-8pt}\times$'' denotes the matrix-vector product.}
    \label{fig:Architecture1}
\end{figure*}

\section{Design of image-captioning-based CAPTCHA}
\label{sec:design}

\subsection{A TPR-capable generation architecture}
\label{subsec:GenArch}
 In this work we propose an approach to network
architecture design we call the \emph{TPR-capable method}. The
architecture we use (see Fig.~\ref{fig:Architecture1}) is designed
so that TPRs could, in theory, be used within the architecture to
perform the target task --- here, generating a caption one word at
a time. Unlike previous work where TPRs are hand-crafted, in our
work, end-to-end deep learning will induce representations which
the architecture can use to generate captions effectively.

As shown in Fig.~\ref{fig:Architecture1}, our proposed system is
denoted by $\mc{N}$.  The input of $\mc{N}$ is an image feature
vector $\mb{v}$ and the output of $\mc{N}$ is a caption. The image
feature vector $\mb{v}$ is extracted from a given image by a
pre-trained CNN. The first part of our system $\mc{N}$ is a
\emph{sentence-encoding subnetwork} $\mc{S}$ which maps $\mb{v}$
to a representation $S$ which will drive the entire
caption-generation process; $S$ contains all the image-specific
information for producing the caption. (We will call a caption a
``sentence''  even though it may in fact be just a noun phrase.)

If $S$ were a TPR of the caption itself, it would be a matrix (or
2-index tensor) $\S$ which is a sum of matrices, each of which
encodes the binding of one word to its role in the sentence
constituting the caption. To serially read out the words encoded
in $\S$, in iteration 1 we would \emph{unbind} the first word from
$\S$, then in iteration 2 the second, and so on. As each word is
generated, $\S$ could update itself, for example, by subtracting
out the contribution made to it by the word just generated;
$\S_{t}$ denotes the value of $\S$ when word $w_{t}$ is generated.
At time step $t$ we would unbind the role $r_{t}$ occupied by word
$w_{t}$ of the caption. So the second part of our system $\mc{N}$
--- the \emph{unbinding subnetwork} $\mc{U}$ --- would generate,
at iteration $t$, the \emph{unbinding vector} $\mb{u}_{t}$. Once
$\mc{U}$ produces the unbinding vector $\mb{u}_{t}$, this vector
would then be applied to $\S$ to extract the symbol $\bf{f}_{t}$
that occupies word $t$'s role; the symbol represented by
$\bf{f}_{t}$ would then be decoded into word $w_{t}$ by the third
part of $\mc{N}$, i.e., the \emph{lexical decoding subnetwork}
$\mc{L}$, which outputs $\mb{x}_{t}$, the 1-hot-vector encoding of
$w_{t}$.

Unbinding in TPR is achieved by the matrix-vector product (see
Appendix~\ref{sec:ReviewTPR}).  So the key operation in generating
$w_{t}$ is thus the unbinding of $r_{t}$ within $\S$, which
amounts to simply:
\begin{eqnarray}
\label{eq:Unbinding} \S_{t} \bf{u}_{t} = \bf{f}_{t}.
\end{eqnarray}
This matrix-vector product is denoted
``$\Box\hspace{-8pt}\times$'' in Fig.~\ref{fig:Architecture1}.

Thus the system $\mc{N}$ of Fig.~\ref{fig:Architecture1} is
TPR-capable. This is what we propose as the Tensor-Product
Generation Network (TPGN) architecture. The learned representation
$\S$ will not be proven to literally be a TPR, but by analyzing
the unbinding vectors $\mb{u}_{t}$ the network learns, we will
gain insight into the process by which the learned matrix $\S$
gives rise to the generated caption.

What type of roles might the unbinding vectors be unbinding? A TPR
for a caption could in principle be built upon \emph{positional
roles}, \emph{syntactic/semantic roles}, or some combination of
the two. In the caption \textbf{\emph{a man standing in a room
with a suitcase}}, the initial \emph{a} and \emph{man} might
respectively occupy the positional roles of
$\textsc{pos(ition)}_{1}$ and  $\textsc{pos}_{2}$; \emph{standing}
might occupy the syntactic role of \textsc{verb}; \emph{in} the
role of \textsc{Spatial-P(reposition)}; while \emph{a room with a
suitcase} might fill a 5-role schema $\textsc{Det(erminer)}_{1}
\textsc{ N(oun)}_{1} \textsc{ P Det}_{2} \textsc{ N}_{2}$. In
fact, there is evidence that our network learns just this kind of
hybrid role decomposition.

What form of information does the sentence-encoding subnetwork
$\mc{S}$ need to encode in $\S$? Continuing with the example of
the previous paragraph, $\S$ needs to be some approximation to the
TPR summing several filler/role binding matrices. In one of these
bindings, a filler vector $\mb{f}_{\textit{a}}$ --- which the
lexical subnetwork $\mc{L}$ will map to the article \emph{a} ---
is bound (via the outer product) to a role vector
$\mb{r}_{\textsc{Pos}_{1}}$ which is the dual of the first
unbinding vector produced by the unbinding subnetwork $\mc{U}$:
$\mb{u}_{\textsc{Pos}_{1}}$. In the first iteration of generation
the model computes $\S_{1} \mb{u}_{\textsc{Pos}_{1}} =
\mb{f}_{\textit{a}}$, which $\mc{L}$ then maps to \emph{a}.
Analogously, another binding approximately contained in $\S_{2}$
is $\mb{f}_{\textit{man}} \mb{r}_{\textsc{Pos}_{2}}^{\top}$. There
are corresponding bindings for the remaining words of the caption;
these employ syntactic/semantic roles. One example is
$\mb{f}_{\textit{standing}} \mb{r}_{V}^{\top}$. At iteration 3,
$\mc{U}$ decides the next word should be a verb, so it generates
the unbinding vector $\mb{u}_{V}$ which when multiplied by the
current output of $\mc{S}$, the matrix $\S_{3}$, yields a filler
vector $\mb{f}_{\textit{standing}}$ which $\mc{L}$ maps to the
output \emph{standing}. $\mc{S}$ decided the caption should deploy
\emph{standing} as a verb and included in $\S$ the binding
$\mb{f}_{\textit{standing}} \mb{r}_{V}^{\top}$. It similarly
decided the caption should deploy \emph{in} as a spatial
preposition, including in $\S$ the binding $\mb{f}_{\textit{in}}
\mb{r}_{\textsc{Spatial-P}}^{\top}$; and so on for the other words
in their respective roles in the caption.

\subsection{CAPTCHA generation method}
\label{subsec:generation}

We first describe how to obtain a set of images as challenges in
our CAPTCHA system. In our proposed image/visual captioning based
CAPTCHA, a TPGN takes an image ${\mathbf I}$ as input and
generates a caption. Then, an evaluator will evaluate the TPGN's
captioning performance by calculating the metric of SPICE
\cite{anderson2016spice} by comparing to human-generated
gold-standard captions. If the SPICE value is less than a
threshold $\gamma_1$, this input image ${\mathbf I}$ can be used
as a challenge in our CAPTCHA; otherwise, this input image
${\mathbf I}$ will not be selected as a challenge and a new image
will be examined.  In this way, we can obtain a set ${\mathbf B}$
of images used as challenges in CAPTCHA.  All images in ${\mathbf
B}$ are difficult to be captioned by a computerized image
captioning system due to their low SPICE values.

Now, we describe our CAPTCHA generation method. In our CAPTCHA, an
image is randomly selected from the set ${\mathbf B}$ and rendered
as a challenge; a tester is asked to give a description of the
image as an answer.  An evaluator calculates a SPICE value for the
answer.  If the SPICE value is greater than a threshold
$\gamma_2$, the answer is considered to be correct and the tester
is deemed to be a human; otherwise, the answer is considered to be
wrong and the tester is deemed to be a computer.

Due to space limitations, the detailed design of our system is
described in Appendix~\ref{sec:System_Description}; and TPR is
reviewed in Appendix~\ref{sec:ReviewTPR}.

\section{Experimental results}
\label{sec:ExperimentalResults}

To evaluate the performance of our proposed architecture, we use
the COCO dataset  \cite{COCO_weblink}. The COCO dataset contains
123,287 images, each of which is annotated with at least 5
captions. We use the same pre-defined splits as
\cite{karpathy2015deep,SCN_CVPR2017}: 113,287 images for training,
5,000 images for validation, and 5,000 images for testing. We use
the same vocabulary as that employed in \cite{SCN_CVPR2017}, which
consists of 8,791 words.


For the CNN of Fig. \ref{fig:Architecture1}, we used ResNet-152
\cite{he2016deep}, pretrained on the ImageNet dataset. The feature
vector ${\mathbf v}$ has 2048 dimensions. Word embedding vectors
in ${\mathbf W}_e$ are downloaded from the web
\cite{Stanford_Glove_weblink}. The model is implemented in
TensorFlow \cite{tensorflow2015-whitepaper} with the default
settings for random initialization and optimization by
backpropagation.

In our experiments, we choose $d=25$ (where $d$ is the dimension
of vector ${\mathbf p}_t$).  The dimension of ${\mathbf S}_t$ is
 $625 \times 625$;
 the vocabulary size $V=8,791$; the dimension of ${\mathbf  u}_t$ and ${\mathbf f}_t$ is
 $d^2=625$.

{\small
\begin{table*}[htb]
  \caption{Performance of the proposed TPGN model on the COCO dataset.}
  \label{table:BLEU}
  \centering
  \begin{tabular}{llllllllll}
    \hline
Methods     & METEOR &BLEU-1 & BLEU-2 & BLEU-3 & BLEU-4 &   CIDEr &SPICE\\
    \hline
NIC \cite{vinyals2015show} & --&  0.666 &0.451 &0.304 &0.203&  --&  --\\
CNN-LSTM &0.238 & 0.698 &0.525 &0.390 &0.292  & 0.889 &  --\\
 TPGN  & \textbf{0.243}& \textbf{0.709} & \textbf{0.539} &  \textbf{0.406} & \textbf{0.305}  &  \textbf{0.909}&  \textbf{0.18}\\
    \hline
  \end{tabular}
\end{table*}
}

The main evaluation results on the MS COCO dataset are reported in
Table~\ref{table:BLEU}. The widely-used BLEU
\cite{papineni2002bleu}, METEOR  \cite{banerjee2005meteor}, CIDEr
\cite{vedantam2015cider}, and SPICE \cite{anderson2016spice}
metrics are reported in our quantitative evaluation of the
performance of the proposed schemes. In evaluation, our baseline
is the widely used CNN-LSTM captioning method originally proposed
in \cite{vinyals2015show}. For comparison, we include results in
that paper in the first line of Table~\ref{table:BLEU}. We also
re-implemented the model using the latest ResNet feature and
report the results in the second line of Table~\ref{table:BLEU}.
Our re-implementation of the CNN-LSTM method matches the
performance reported in \cite{SCN_CVPR2017}, showing that the
baseline is a state-of-the-art implementation. As shown in
Table~\ref{table:BLEU}, compared to the CNN-LSTM baseline, the
proposed TPGN significantly outperforms the benchmark schemes in
all metrics across the board. The improvement in BLEU-$n$ is
greater for greater $n$; TPGN particularly improves generation of
longer subsequences. The results clearly attest to the
effectiveness of the TPGN architecture.

Now, we address the issue of how to determine the two thresholds
$\gamma_1$ and $\gamma_2$ in our CAPTCHA system. We set
$\gamma_1=0.04$, which is about 80\% less than the SPICE metric
obtained by TPGN.  In this way, the image set ${\mathbf B}$ only
contains images that are most difficult to be captioned by a
computer.

We run a trained TPGN with input images from COCO test dataset,
and select ten images from COCO test dataset, whose SPICE metrics
are less than $\gamma_1$.  Then we use Amazon Mechanical Turk
system to generate captions for these ten images by humans.  We
observe that the SPICE metric obtained by a caption generated by a
human is always larger than 0.3; hence, we set $\gamma_2=0.3$,
which is much larger than the SPICE metric achievable by any
existing computerized image captioning system \cite{COCO_weblink}.


\section{Conclusion}
\label{sec:Conclusion}

In this paper, we proposed a new Tensor Product Generation Network
(TPGN) for image/visual-captioning-based CAPTCHAs. The model has a
novel architecture based on a rationale derived from the use of
Tensor Product Representations for encoding and processing
symbolic structure through neural network computation. In
evaluation, we tested the proposed model on captioning with the MS
COCO dataset, a large-scale image-captioning benchmark. Compared
to widely adopted LSTM-based models, the proposed TPGN gives
significant improvements on all major metrics including METEOR,
BLEU, CIDEr, and SPICE.  Our findings in this paper show great
promise of TPRs in CAPTCHA. In the future, we will explore
extending TPR to a variety of other NLP tasks and spam email
detection.

\section*{Appendix}
\appendix
\section{Review of tensor product representation}
\label{sec:ReviewTPR}

Tensor product representation (TPR) is a general framework for
embedding a space of symbol structures $\mathfrak{S}$ into a
vector space. This embedding enables neural network operations to
perform symbolic computation, including computations that provide
considerable power to symbolic NLP systems
\cite{smolensky2006harmonic,smolensky2012symbolic}. Motivated by
these successful examples, we are inspired to extend the TPR to
the challenging task of learning image captioning. And as a
by-product, the symbolic character of TPRs makes them amenable to
conceptual interpretation in a way that standard learned neural
network representations are not.

A particular TPR embedding is based in a \emph{filler/role
decomposition} of $\mathfrak{S}$ . A relevant example is when
$\mathfrak{S}$ is the set of strings over an alphabet $\{ \mt{a},
\mt{b}, \ldots\}$. One filler/role decomposition deploys the
\emph{positional roles} $\{ r_{k} \}, k \in \mathbb{N}$, where the
\emph{filler/role binding} $\mt{a}/r_{k}$ assigns the `filler'
(symbol) $\mt{a}$ to the $k^{th}$ position in the string. A string
such as $\mt{abc}$ is uniquely determined by its filler/role
bindings, which comprise the (unordered) set
$\mathfrak{B}(\mt{abc}) = \{ \mt{b}/r_{2}, \mt{a}/r_{1},
\mt{c}/r_{3} \}$. Reifying the notion \emph{role} in this way is
key to TPR's ability to encode complex symbol \emph{structures}.

Given a selected filler/role decomposition of the symbol space, a
particular TPR is determined by an embedding that assigns to each
filler a vector in a vector space $V_{F} \cong
\mathbb{R}^{d_{F}}$, and a second embedding that assigns to each
role a vector in a space $V_{R}\cong \mathbb{R}^{d_{R}}$. The
vector embedding a symbol $\mt{a}$ is denoted by $\mb{f_{\mt{a}}}$
and is called a \emph{filler vector}; the vector embedding a role
$r_{k}$ is $\mb{r_{k}}$ and called a \emph{role vector}. The TPR
for $\mt{abc}$ is then the following 2-index tensor in $V_{F}
\otimes V_{R} \cong \mathbb{R}^{d_{F}\times d_{R}}$:
\begin{eqnarray}
\bf{S}_{\mt{\mt{abc}}} = \mb{f_{\mt{b}}} \otimes \mb{r_{2}} +
\mb{f_{\mt{a}}} \otimes \mb{r_{1}} + \mb{f_{\mt{c}}} \otimes
\mb{r_{3}}, \label{eqn:TPRdefinition1}
\end{eqnarray}
where $\otimes$ denotes the tensor product. The tensor product is
a generalization of the vector outer product that is recursive;
recursion is exploited in TPRs for, e.g., the distributed
representation of trees, the neural encoding of formal grammars in
connection weights, and the theory of neural computation of
recursive symbolic functions. Here, however, it suffices to use
the outer product; using matrix notation we can write
(\ref{eqn:TPRdefinition1}) as:
\begin{eqnarray}
\bf{S}_{\mt{abc}} = \mb{f_{\mt{b}}} \mb{r_{2}}^{\!\!\top} +
\mb{f_{\mt{a}}} \mb{r_{1}}^{\!\!\top} + \mb{f_{\mt{c}}}
\mb{r_{3}}^{\!\!\top}. \label{eqn:TPRdefinition2}
\end{eqnarray}
Generally, the embedding of any symbol structure $\mt{S} \in
\mathfrak{S}$ is $\sum \{ \mb{f}_{i} \otimes \mb{r}_{i} \mid
\mt{f}_{i}/r_{i} \in \mathfrak{B}(\mt{S}) \}$; here: $\sum \{
\mb{f}_{i} \mb{r}_{i}^{\top} \mid \mt{f}_{i}/r_{i} \in
\mathfrak{B}(\mt{S}) \}$
\cite{smolensky1990tensor,smolensky2006harmonic}.

A key operation on TPRs, central to the work presented here, is
\emph{unbinding}, which undoes binding. Given the TPR in
(\ref{eqn:TPRdefinition2}), for example, we can unbind
$\mb{r_{2}}$ to get $\mb{f_{\mt{b}}}$; this is achieved simply by
$\mb{f_{\mt{b}}} = \mb{S_{\mt{abc}}} \mb{u_{2}}$. Here
$\mb{u_{2}}$ is \emph{the unbinding vector dual to} the binding
vector $\mb{r_{2}}$. To make such exact unbinding possible, the
role vectors should be chosen to be linearly independent. (In that
case the unbinding vectors are the rows of the inverse of the
matrix containing the binding vectors as  columns, so that
$\mb{r_{2} \cdot u_{2}} = 1$  while $\mb{r_{k} \cdot u_{2}} = 0$
for all other role vectors $\mb{r_{k} \neq r_{2}}$; this entails
that $\mb{S_{\mt{abc}}} \mb{u_{2}} = \mb{b}$, the filler vector
bound to $\mb{r_{2}}$. Replacing the matrix inverse with the
pseudo-inverse allows approximate unbinding when the role vectors
are not linearly independent).

\section{System Description}
\label{sec:System_Description}

\begin{figure}[htb]
    \centering
    \includegraphics[width=0.8\textwidth]{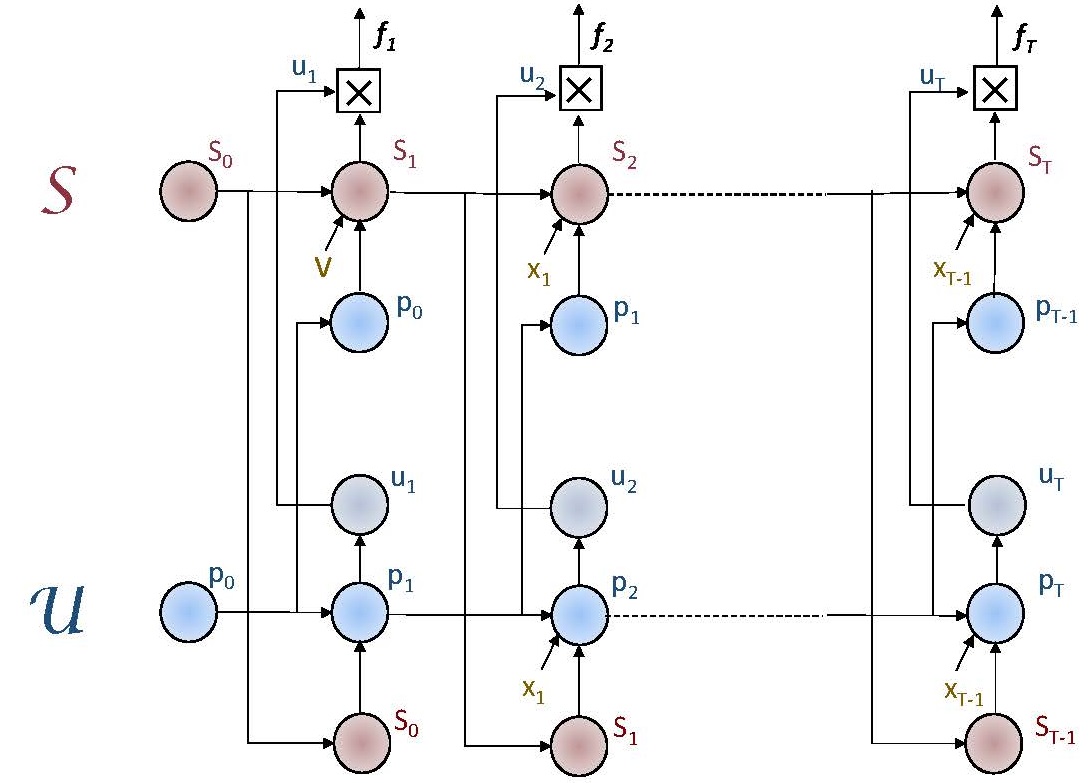}
    \caption{The sentence-encoding subnet $\mc{S}$ and the unbinding subnet $\mc{U}$ are inter-connected LSTMs; $\mb{v}$ encodes the visual input while the $\mb{x}_{t}$ encode the words of the output caption.
    }
    \label{fig:Architecture2}
\end{figure}

The unbinding subnetwork $\mc{U}$ and the sentence-encoding
network $\mc{S}$ of Fig. \ref{fig:Architecture1} are each
implemented as (1-layer, 1-directional) LSTMs (see Fig.
\ref{fig:Architecture2}); the lexical subnetwork $\mc{L}$ is
implemented as a linear transformation followed by a softmax
operation.
In the equations below, the LSTM variables internal to the
$\mc{S}$ subnet are indexed by 1 (e.g., the forget-, input-, and
output-gates are respectively $\mb{\hat{f}}_{1}, \mb{\hat{i}}_{1},
\mb{\hat{o}}_{1}$) while those of the unbinding subnet $\mc{U}$
are indexed by 2.

Thus the state updating equations for $\mc{S}$ are, for
$t=1,\cdots,T$ = caption length: {\small
\begin{eqnarray}
{\mathbf {\hat{f}}}_{1,t}&=&\sigma_g({\mathbf W}_{1,f} {\mathbf
p}_{t-1}-{\mathbf D}_{1,f}{\mathbf W}_e{\mathbf x}_{t-1}+{\mathbf
U}_{1,f}{\mathbf {\hat{S}}}_{t-1})\label{eq:TPR-LSTM1_1}\\
{\mathbf {\hat{i}}}_{1,t}&=&\sigma_g({\mathbf W}_{1,i} {\mathbf
p}_{t-1}-{\mathbf D}_{1,i}{\mathbf W}_e {\mathbf x}_{t-1}+{\mathbf
U}_{1,i}{\mathbf {\hat{S}}}_{t-1})\\
{\mathbf {\hat{o}}}_{1,t}&=&\sigma_g({\mathbf W}_{1,o} {\mathbf
p}_{t-1}-{\mathbf D}_{1,o}{\mathbf W}_e {\mathbf x}_{t-1}+{\mathbf
U}_{1,o}{\mathbf {\hat{S}}}_{t-1})\\
{\mathbf g}_{1,t}&=&\sigma_h({\mathbf W}_{1,c} {\mathbf
p}_{t-1}-{\mathbf D}_{1,c}{\mathbf W}_e {\mathbf x}_{t-1}+{\mathbf
U}_{1,c}{\mathbf {\hat{S}}}_{t-1})\\
{\mathbf c}_{1,t}&=& {\mathbf {\hat{f}}}_{1,t} \odot {\mathbf
c}_{1,t-1}+
{\mathbf {\hat{i}}}_{1,t} \odot {\mathbf g}_{1,t}\\
 {\mathbf {\hat{S}}}_t&=& {\mathbf {\hat{o}}}_{1,t} \odot \sigma_h({\mathbf c}_{1,t})
\label{eq:TPR-LSTM1_5}
\end{eqnarray}    }
where ${\mathbf {\hat{f}}}_{1,t}$, ${\mathbf {\hat{i}}}_{1,t}$,
${\mathbf {\hat{o}}}_{1,t}$, ${\mathbf g}_{1,t}$, ${\mathbf
c}_{1,t}$,
${\mathbf {\hat{S}}}_t\in \mathbb{R}^{d\times d}$, ${\mathbf p}_t$
$\in \mathbb{R}^{d}$,
$\sigma_g(\cdot)$ is the (element-wise) logistic sigmoid function;
$\sigma_h(\cdot)$ is the hyperbolic tangent function; the operator
$\odot$ denotes the Hadamard (element-wise) product; ${\mathbf
W}_{1,f},{\mathbf W}_{1,i},{\mathbf W}_{1,o},{\mathbf W}_{1,c} \in
\mathbb{R}^{d\times d \times d}$, ${\mathbf D}_{1,f}$, $ {\mathbf
D}_{1,i}$, $ {\mathbf D}_{1,o}$, $ {\mathbf D}_{1,c}$ $\in$
$\mathbb{R}^{d\times d \times d}$, ${\mathbf U}_{1,f}$, $ {\mathbf
U}_{1,i}$, $ {\mathbf U}_{1,o}$, $ {\mathbf U}_{1,c}$ $\in$
$\mathbb{R}^{d\times d \times d\times d}$. For clarity, biases ---
included throughout the model --- are omitted from all equations
in this paper. The initial state ${\mathbf {\hat{S}}}_0$ is
initialized by:
\begin{equation}
{\mathbf {\hat{S}}}_0={\mathbf C}_s ({\mathbf v}-\bar{{\mathbf
v}})
 \label{eq:TPR-RNN1_4}
\end{equation}
where ${\mathbf v} \in \mathbb{R}^{2048}$ is the vector of visual
features extracted from the current image by ResNet
\cite{SCN_CVPR2017} and $\bar{{\mathbf v}}$ is the mean of all
such vectors;
 ${\mathbf C}_s \in
\mathbb{R}^{d\times d \times 2048}$. On the output side, ${\mathbf
x}_t \in \mathbb{R}^{V}$ is a 1-hot vector with dimension equal to
the size of the caption vocabulary, $V$, and ${\mathbf W}_e \in
\mathbb{R}^{d\times V}$ is a word embedding matrix, the $i$-th
column of which is the embedding vector of the $i$-th word in the
vocabulary; it is obtained by the Stanford GLoVe algorithm with
zero mean \cite{Stanford_Glove_weblink}. ${\mathbf x}_0$ is
initialized as the one-hot vector corresponding to a
 ``start-of-sentence'' symbol.

For $\mc{U}$ in Fig.~\ref{fig:Architecture1}, the state updating
equations are:{\small
\begin{eqnarray}
{\mathbf {\hat{f}}}_{2,t}&=&\sigma_g({\mathbf
{\hat{S}}}_{t-1}{\mathbf w}_{2,f}-{\mathbf D}_{2,f}{\mathbf W}_e
{\mathbf
x}_{t-1}+{\mathbf U}_{2,f}{\mathbf p}_{t-1})\label{eq:TPR-LSTM2_1}\\
{\mathbf {\hat{i}}}_{2,t}&=&\sigma_g({\mathbf
{\hat{S}}}_{t-1}{\mathbf w}_{2,i}-{\mathbf D}_{2,i}{\mathbf W}_e
{\mathbf x}_{t-1}+{\mathbf
U}_{2,i}{\mathbf p}_{t-1})\\
{\mathbf {\hat{o}}}_{2,t}&=&\sigma_g({\mathbf
{\hat{S}}}_{t-1}{\mathbf w}_{2,o}-{\mathbf D}_{2,o}{\mathbf W}_e
{\mathbf x}_{t-1}+{\mathbf
U}_{2,o}{\mathbf p}_{t-1})\\
{\mathbf g}_{2,t}&=&\sigma_h({\mathbf {\hat{S}}}_{t-1}{\mathbf
w}_{2,c}-{\mathbf D}_{2,c}{\mathbf W}_e {\mathbf x}_{t-1}+{\mathbf
U}_{2,c}{\mathbf p}_{t-1})\\
 {\mathbf c}_{2,t}&=& {\mathbf {\hat{f}}}_{2,t} \odot
{\mathbf c}_{2,t-1}+ {\mathbf {\hat{i}}}_{2,t} \odot {\mathbf g}_{2,t}\\
{\mathbf p}_t&=&{\mathbf {\hat{o}}}_{2,t} \odot \sigma_h({\mathbf
c}_{2,t})
 \label{eq:TPR-LSTM2_5}
\end{eqnarray}   }
where ${\mathbf w}_{2,f}, {\mathbf w}_{2,i}, {\mathbf w}_{2,o},
{\mathbf w}_{2,c} \in \mathbb{R}^{d}$, ${\mathbf D}_{2,f},{\mathbf
D}_{2,i},{\mathbf D}_{2,o},{\mathbf D}_{2,c}$ $\in$ $\mathbb{R}^{d
\times d}$, and ${\mathbf U}_{2,f}, {\mathbf U}_{2,i}, {\mathbf
U}_{2,o}, {\mathbf U}_{2,c}$ $\in$ $\mathbb{R}^{d\times d}$. The
initial state ${\mathbf p}_0$ is the zero vector.

The dimensionality of the crucial vectors shown in Fig.
\ref{fig:Architecture1}, $\mb{u}_{t}$ and $\mb{f}_{t}$, is
increased from $d \times 1$ to $d^{2} \times 1$ as follows. A
block-diagonal $d^{2} \times d^{2}$ matrix $\mb{S}_{t}$ is created
by placing $d$ copies of the $d \times d$ matrix
$\mb{\hat{S}}_{t}$ as blocks along the principal diagonal. This
matrix is the output of the sentence-encoding subnetwork $\mc{S}$.
Now, following Eq. \eqref{eq:Unbinding}, the  `filler vector'
${\mathbf f}_t\in \mathbb{R}^{d^{2}}$ --- `unbound' from the
sentence representation $\S_{t}$ with the `unbinding vector'
$\mb{u}_{t}$ --- is obtained by Eq. \eqref{eq:TPR-RNN1f}.
\begin{equation}
{\mathbf f}_t= {\mathbf S}_t  {\mathbf u}_t
 \label{eq:TPR-RNN1f}
\end{equation}
Here ${\mathbf u}_t\in \mathbb{R}^{d^{2}}$, the output of the
unbinding subnetwork $\mc{U}$, is computed as in Eq.
\eqref{eq:TPR-RNN2u}, where ${\mathbf W}_u \in
\mathbb{R}^{d^{2}\times d}$ is $\mc{U}$'s output weight matrix.
\begin{equation}
{\mathbf u}_t=\sigma_h({\mathbf W}_u{\mathbf p}_t)
 \label{eq:TPR-RNN2u}
\end{equation}

Finally, the lexical subnetwork $\mc{L}$ produces a decoded word
${\mathbf x}_t\in \mathbb{R}^{V}$ by
\begin{equation}
{\mathbf x}_t=\sigma_s({\mathbf W}_x{\mathbf f}_t)
 \label{eq:TPR-FFNN}
\end{equation}
where $\sigma_s(\cdot)$ is the softmax function and ${\mathbf W}_x
\in \mathbb{R}^{V\times d^2}$ is the overall output weight matrix.
Since ${\mathbf W}_x$ plays the role of a word de-embedding
matrix, we can set
\begin{equation}
{\mathbf W}_x=({\mathbf W}_e)^\top
 \label{eq:embeddingVector2}
\end{equation}
where ${\mathbf W}_e$ is the word-embedding matrix. Since
${\mathbf W}_e$ is pre-defined, we directly set ${\mathbf W}_x$ by
Eq.~\eqref{eq:embeddingVector2} without training $\mc{L}$ through
Eq.~\eqref{eq:TPR-FFNN}. Note that $\mc{S}$ and $\mc{U}$ are
learned jointly through the end-to-end training.

\begin{figure*}[tbh]
    \centering
    \includegraphics[width=\textwidth]{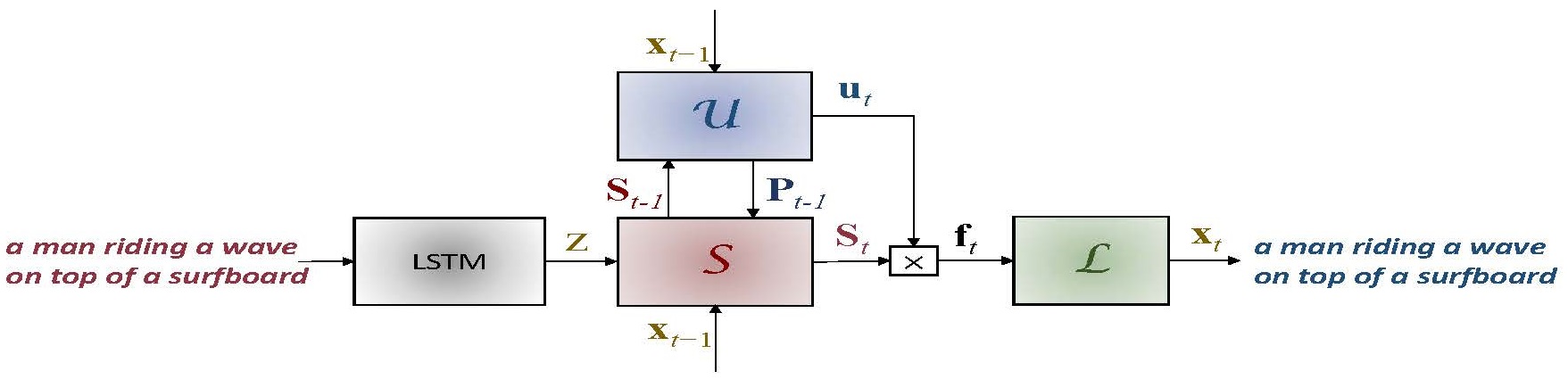}
    \caption{Pre-training of TPGN.}
    \label{fig:pretraining}
\end{figure*}

Fig.~\ref{fig:pretraining} shows a pre-training method for
initializing TPGN. During the pre-training phase, there is no
image input, i.e., image feature vector ${\mathbf v}=0$. In
Fig.~\ref{fig:pretraining}, at time $t=-T+1$, the LSTM module
takes a sentence of length $T$ as input and outputs a vector
${\mathbf z}$ (${\mathbf z}\in \mathbb{R}^{d^2}$) at time $t=0$.
That is, the LSTM converts a sentence into ${\mathbf z}$, which is
the input of TPGN.  We use end-to-end training to train the whole
system shown in Fig.~\ref{fig:pretraining}. After finishing
pre-training, we let ${\mathbf z}=0$ and use images as input to
train the TPGN in Fig.~\ref{fig:Architecture1}, initialized by the
pretrained parameter values.

\section{Related work}
\label{sec:RelatedWork}

Most existing DL-based image captioning methods
\cite{mao2015deep,vinyals2015show,devlin2015language,chen2015mind,donahue2015long,karpathy2015deep,kiros2014multimodal,kiros2014unifying}
involve two phases/modules: 1) image analysis, typically by a
Convolutional Neural Network (CNN), and 2) a language model for
caption generation
(\cite{fang2015captions}).  
The CNN module takes an image as input and outputs an image
feature vector or a list of detected words with their
probabilities. The language model is used to create a sentence
(caption) out of the detected words or the image feature vector
produced by the CNN.

There are mainly two approaches to natural language generation in
image captioning. The first approach takes the words detected by a
CNN as input, and uses a probabilistic model, such as a maximum
entropy (ME) language model, to arrange the detected words into a
sentence. The second approach takes the penultimate activation
layer of the CNN as input to a Recurrent Neural Network (RNN),
which generates a sequence of words (the caption)
\cite{vinyals2015show}.

The work reported here follows the latter approach, adopting a CNN
+ RNN-generator architecture.  Specifically, instead of using a
conventional RNN, we propose a recurrent network that has
substructure derived from the general GSC architecture: one
recurrent subnetwork holds an encoding $\S$
--- which is treated as an approximation of a TPR
--- of the words yet to be produced, while another recurrent
subnetwork generates a sequence of vectors that is treated as a
sequence of roles to be unbound from $\S$, in effect, reading out
a word at a time from $\S$. Examining how the model deploys these
roles allows us to interpret them in terms of grammatical
categories; roughly speaking, a sequence of categories is
generated and the words stored in $\S$ are retrieved and spelled
out via their categories.

\end{document}